\documentclass[11pt,a4paper]{article}

\usepackage[bottom]{footmisc}                                       \makeatletter
\def\blfootnote{\xdef\@thefnmark{}\@footnotetext}
\makeatother

\usepackage[hyperref]{emnlp-ijcnlp-2019}
\usepackage{times}
\usepackage{latexsym}
\usepackage{url}

\aclfinalcopy 

\usepackage{url}
\usepackage{graphicx}
\usepackage{booktabs}
\usepackage{bm}
\usepackage{amsfonts}
\usepackage{amsmath}
\usepackage{multirow}
\usepackage{xspace}
\usepackage{linguex}

\usepackage{siunitx}
\usepackage[toc,page]{appendix}

\newcommand{\bl}[1]{\textcolor{blue}{#1}}
\newcommand{\bfbl}[1]{\textcolor{blue}{\textbf{#1}}}
\newcommand{\rd}[1]{\textcolor{red}{#1}}

\newcommand\bertbase{BERT$_{\small \textsc{BASE}}$\xspace}
\newcommand\bertlarge{BERT$_{\small \textsc{LARGE}}$\xspace}

\title{Improving Natural Language Inference with a Pretrained Parser}

\author{
  Deric Pang$^\star$ \quad
  Lucy H. Lin$^\star$ \quad
  Noah A. Smith$^{\star\dagger}$ \\
  $^\star$Paul G. Allen School of Computer Science \& Engineering,
    University of Washington \\
  $^\dagger$Allen Institute for Artificial Intelligence \\
  {\tt \{dericp,lucylin,nasmith\}@cs.washington.edu} \\
}

\date{}

\begin{document}
\maketitle
\begin{abstract}



We introduce a novel approach to incorporate syntax into natural language
inference (NLI) models. Our method uses contextual token-level vector
representations from a pretrained dependency parser.
Like other contextual embedders, our method is broadly
applicable to any neural model.  We experiment with four strong NLI models
(decomposable attention model, ESIM, BERT, and MT-DNN), and show consistent
benefit to accuracy across three NLI benchmarks.\blfootnote{Our code is available
at \url{https://github.com/dericp/syntactic-entailment}.}

\end{abstract}

\section{Introduction}
We consider natural language inference (NLI)
tasks in which the semantic
relationship between two sentences is classified as entailment, contradiction,
or neither.
Our focus is on the use of syntactic representations of the
sentences, given (1) longstanding linguistic theory that posits a close
relationship between syntax and semantics \citep[e.g.,][]{montague1970universal,steedman2011combinatory} and (2) major
advances in syntactic parsing accuracy
(e.g., \citealp{kiperwasser2016simple,dozat2016deep}).

Building on the recent success of transferring contextual word representations
learned by large-data language modeling or translation to other tasks
\citep{mccann2017learned,peters2018deep,radford2018improving,Devlin:19}, we introduce two general methods for transferring
word representations from a \emph{parser} to an NLI model (\S\ref{sec:model}) --- through input into a model's final feedforward (classification) layer, or through a model's attention mechanism.

We see widespread gains when applying these methods to
four strong NLI models (decomposable attention model, ESIM, BERT, and MT-DNN)
on the SNLI, MNLI, and SciTail datasets (\S\ref{sec:experiments}).
We also provide analysis for how including syntactic representations
from a parser is helpful. In particular, we probe our models with the Heuristic Analysis for NLI Systems (HANS)
dataset and show that when our methods improve NLI test accuracy, performance on this evaluation dataset
improves as well (\S\ref{sec:analysis}).

\section{Model}
\label{sec:model}
\begin{figure}[t]
    \centering
    \includegraphics[width=\linewidth]{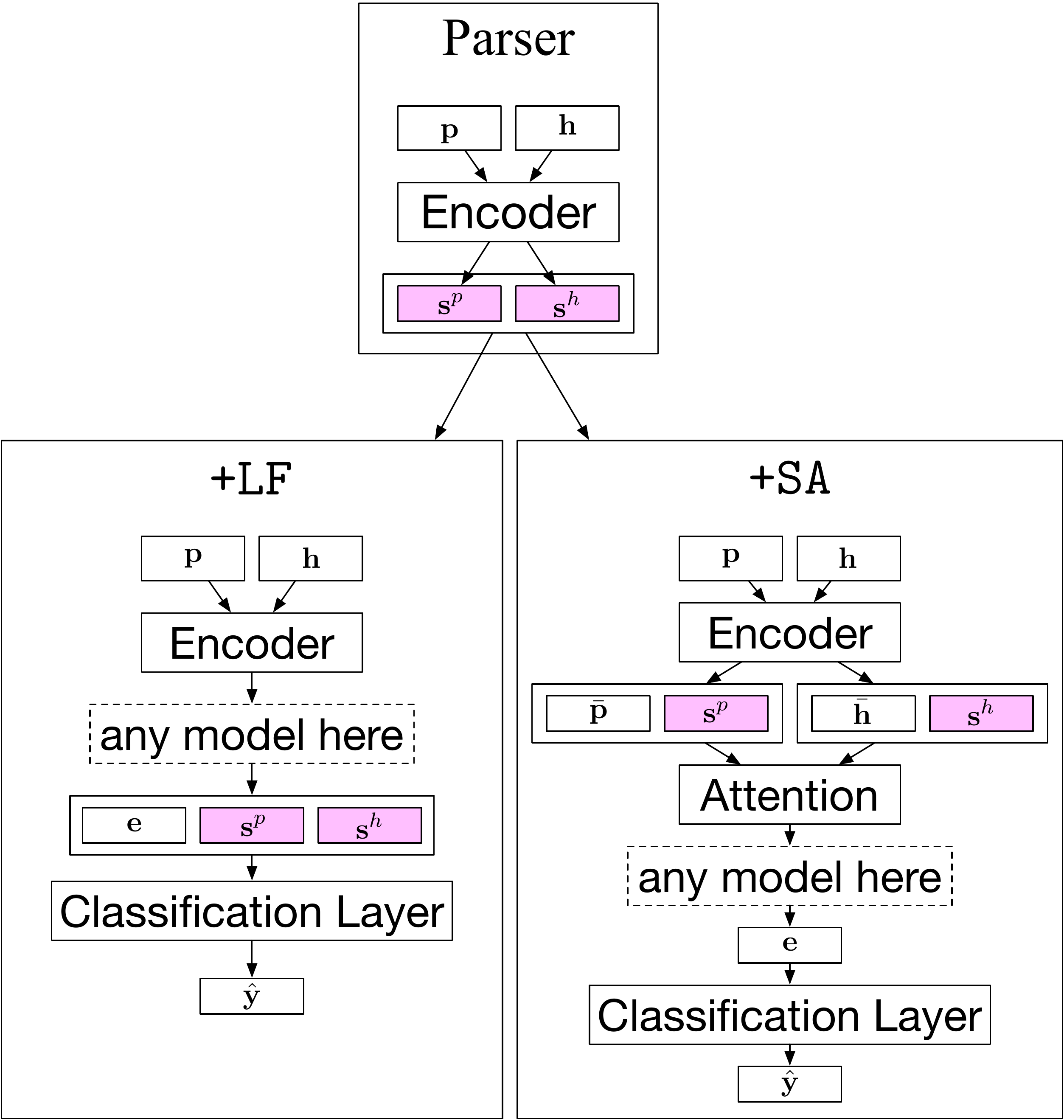}
    \caption{Our methods for incorporating syntactic representations, \texttt{+LF} (left) and \texttt{+SA} (right). $\bm{p}$ and $\bm{h}$ are the premise and hypothesis.}
\label{fig:syntactic-attention-model}
\end{figure}

In NLI tasks, given a \emph{premise} and a \emph{hypothesis}, a system
is asked to
determine whether the hypothesis is entailed by the premise, contradicts the
premise, or is neutral to the premise.

Let the premise be represented as a sequence of token embeddings, $\langle
\bm{p}_1, \dots, \bm{p}_{\ell_p} \rangle$, and likewise for the hypothesis,
$\langle \bm{h}_1, \dots, \bm{h}_{\ell_h} \rangle$. An NLI model will take
these sequences as input and apply some series of transformations to create an
output vector $\bm{e}$. From there, a final feedforward layer ($H$) is applied
to $\bm{e}$ to compute the final prediction:
\begin{align}\label{eq:predict}
    \hat{\bm{y}} = H(\bm{e}).
\end{align}

\subsection{Extracting Syntactic Word Representations}

Given a neural syntax parser with an encoder (e.g., LSTM or transformer)
that encodes a representation of the input sequence from
which a parse is later computed, we can obtain contextual token-level vector
representations of the premise and hypothesis,
$\langle \bm{s}^p_1, \dots, \bm{s}^p_{\ell_p} \rangle$ and
$\langle \bm{s}^h_1, \dots, \bm{s}^h_{\ell_h} \rangle$,
which we call ``syntactic word representations'' (SWRs).
In our method, $\bm{s}^p$ and $\bm{s}^h$ are the hidden states of the
encoder when (separately) parsing the premise and hypothesis.

\subsection{Late Fusion of SWRs}

The simplest way to incorporate the extracted SWRs is to
attach them to $\bm{e}$, the input to the final feedforward layer. More
precisely, we concatenate the final contextual representations from the parser
($\bm{s}^p_{\ell_p}$, $\bm{s}^h_{\ell_h}$) to $\bm{e}$. The prediction step in
Equation~\ref{eq:predict} then becomes:
\begin{align}
    \hat{\bm{y}} = H([\bm{e}, \bm{s}^p_{\ell_p}, \bm{s}^h_{\ell_h}])
\end{align}
This approach is very general, since many neural models use
a final feedforward classification layer, and adds a minimal number of parameters.
We will refer to this variant of our method as Late Fusion (\texttt{+LF};
Fig.~\ref{fig:syntactic-attention-model}, left).

\subsection{Using SWRs to Attend}

A natural place to include SWRs in models that use
attention is in conjunction with the attention weights; that is, perhaps syntax
would be helpful for guiding soft alignments between subphrases.
Calculating attention is often formulated as the dot product between two
sequence representations. In NLI, this will be an
encoded representation of the premise and hypothesis ($\bar{\bm{p}}, \bar{\bm{h}})$.
Attention between the $i$th and $j$th tokens
in the premise and hypothesis is calculated with:
\begin{align} \label{eq:attention}
    a_{ij} := \bar{\bm{p}_i}^\mathsf{T} \bar{\bm{h}_j}
\end{align}
and used downstream to ultimately compute $\bm{e}$.

To augment the attention calculation with SWRs, we modify
Equation~\ref{eq:attention} to be:
\begin{align}
    a_{ij} := [\bar{\bm{p}_i}, \bm{s}^p_i]^\mathsf{T} [\bar{\bm{h}_j}, \bm{s}^h_j].
\end{align}

Although the model will not learn any weights
directly on the SWRs, this method works well in practice and does
not add any additional parameters.
We will refer to this attention-level variant of our method as
Syntactic Attention
(\texttt{+SA}; Fig.~\ref{fig:syntactic-attention-model}, right).


\section{Experiments}
\label{sec:experiments}


We experiment on three NLI datasets:
SNLI \citep{bowman2015large}, MNLI
\citep{williams2017broad}, and SciTail \citep{Khot2018-th}.
We extend two models with both \texttt{+LF} and \texttt{+SA},
decomposable attention model and ESIM, and extend two
recent or current state-of-the-art models with \texttt{+LF}, BERT
and MT-DNN.\footnote{Multi-head self-attention (in BERT and MT-DNN) opens up many possibilities for \texttt{+SA} variants to be explored in the future.}

\subsection{Datasets}

We briefly summarize the datasets below.

\paragraph{SNLI.} This dataset consists of 570k human-created and annotated
sentence pairs; the premise sentences are drawn from the Flickr 30k corpus. 

\paragraph{MNLI.} This dataset (433k pairs) was created using a similar
methodology to SNLI, but with text drawn from ten domains. The development and test sets are separated into two
categories: matched (MNLI-m; in-domain) and mismatched (MNLI-mm; cross-domain). 

\paragraph{SciTail.} This dataset (27k pairs) is derived from
science multiple-choice questions. Unlike SNLI and MNLI, the text was not
deliberately generated for the corpus and there is no contradiction label.

\subsection{Baseline Models}

We briefly summarize the baseline models below.

\paragraph{Decomposable attention model.} The Decomposable Attention model (DA;
\citealp{parikh2016decomposable}) has three steps: attend between the premise
and hypothesis, compare aligned subphrases,
and aggregate the comparisons to make a final prediction.  Notably, DA does not
use word order information.





\paragraph{ESIM.} The Enhanced Sequence Inference Model (ESIM;
\citealp{chen2016enhanced}) encodes sentences with a variant of a bidirectional
LSTM. We use the sequential variant of ESIM without syntactic parsing
information in tree LSTMs.

\begin{table*}[tb]
\centering
{\small
\begin{tabular}{lcccccccc}
\toprule
               & \multicolumn{2}{c}{SNLI} & \multicolumn{2}{c}{SciTail} & \multicolumn{2}{c}{MNLI-m} & \multicolumn{2}{c}{MNLI-mm} \\
                 \cmidrule(lr){2-3}\cmidrule(lr){4-5}\cmidrule(lr){6-7}\cmidrule(lr){8-9}
Model          & dev.         & test        & dev.         & test        & dev.         & test        & dev.         & test        \\
\midrule                                   
DA             & 81.7        & 82.1        & 78.3        & 75.1        & 68.6        & 68.6        & 69.4        & 68.9        \\
DA\texttt{+LF} & \bfbl{84.9} & \bfbl{84.8} & \bl{79.3}   & \bl{78.0}   & \bfbl{71.6} & \bfbl{71.6} & \bfbl{72.4} & \bfbl{71.0} \\
DA\texttt{+SA} & \bl{83.1}   & \bl{83.2}   & \bfbl{82.8} & \bfbl{78.2} & \bl{69.8}   & \bl{69.4}   & \bl{71.0}   & \bl{69.6}   \\
\midrule                                   
ESIM             & \textbf{88.9} & 87.9        & 80.7        & 76.2        & \textbf{77.6} & \textbf{77.6} & 77.4        & \textbf{76.2} \\
ESIM\texttt{+LF} & \textbf{88.9} & \rd{87.7}   & \bl{83.7}   & \bl{78.0}   & \rd{77.2}     & \rd{77.3}     & \bfbl{77.7} & \rd{76.1}     \\
ESIM\texttt{+SA} & \rd{88.4}     & \bfbl{88.1} & \bfbl{84.9} & \bfbl{81.3} & \rd{76.9}     & \rd{77.1}     & \bl{77.5}   & \rd{75.8}     \\
\midrule                                   
BERT             & 90.5        & 89.9        & \textbf{94.2} & 90.8        & 84.3        & 84.7        & \textbf{84.7} & \textbf{83.5} \\
BERT\texttt{+LF} & \bfbl{90.6} & \bfbl{90.5} & \rd{93.9}     & \bfbl{92.8} & \bfbl{84.7} & \bfbl{84.9} & \textbf{84.7}   & \rd{83.3}     \\
\midrule                                   
MT-DNN             & \textbf{91.4} & \textbf{91.1} & \textbf{95.7} & 94.0          & 84.2          & 84.1        & \textbf{84.7} & 83.3        \\
MT-DNN\texttt{+LF} & \rd{91.3}     & \textbf{91.1} & \rd{95.1}     & \bfbl{94.3}   & \bfbl{84.3}   & \bfbl{84.5} & \rd{84.6}     & \bfbl{83.8} \\
\bottomrule
\end{tabular}
}
\caption{NLI dev./test accuracies. \textbf{Bold} is the top value on one dataset/architecture pair.  \bl{Blue} numbers represent accuracy increases and \rd{red} numbers decreases, both relative to the syntax-free baseline of the same architecture.}
\label{table:nli}
\end{table*}

\paragraph{BERT.} Bidirectional Encoder Representations from Transformers
(BERT; \citealp{Devlin:19}) is a transformer model pretrained on massive
amounts of text. It can be fine-tuned to a specific dataset or task.

\paragraph{MT-DNN.} The Multi-Task Deep Neural Network (MT-DNN; \citealp{liu2019multi}) is a carefully fine-tuned
BERT model multi-tasked on the nine GLUE
tasks \citep{wang2018glue}.  Like BERT, it can be fine-tuned to a specific
dataset or task. At publication time, MT-DNN was state of the art on SNLI, SciTail, MNLI, and GLUE.

\subsection{Implementation Details}

In all of our experiments, we use the deep biaffine attention model for
dependency parsing by \citet{dozat2016deep};\footnote{In preliminary
experiments, we also explored the use of internal representations from a
constituency parser; these experiments suggested that dependency parser
representations worked better for our purposes.} in particular, we use the
pretrained version from AllenNLP \citep{Gardner2017AllenNLP} trained on the English
Penn Treebank \citep{marcus1993building} with Universal Dependencies \citep{nivre2016universal}.
The parser's parameters are frozen in all of our
experiments.

For DA and ESIM, we use the implementations in
AllenNLP;
for BERT, we use the pretrained
uncased \bertbase from Hugging Face's PyTorch
implementation;\footnote{\url{https://github.com/huggingface/pytorch-transformers}}
and for MT-DNN we use the pretrained uncased \bertbase\footnote{Note that prior state-of-the-art
MT-DNN results on GLUE (and as a result, MNLI) used
\bertlarge, which we do not experiment with.} MT-DNN model from the
original authors.\footnote{\url{https://github.com/namisan/mt-dnn}}

We perform random hyperparameter search for DA and ESIM and use hyperparameters
reported in the original papers for BERT and MT-DNN.
More details about our hyperparameters and tuning procedure can be found in
Appendix A.

\subsection{Results}

Our experimental results are in Table~\ref{table:nli}. We find that
adding representations from the syntax parser (\texttt{+LF}, \texttt{+SA})
always improves test accuracy on SNLI and SciTail across all baseline models. On
MNLI, the results are mixed---DA, BERT, and MT-DNN improve on the matched
subset, and only DA and MT-DNN improve on the mismatched subset.

SWRs always improve DA regardless of the method (\texttt{+LR}, \texttt{+SA})
or dataset.
Despite the massive pretraining of BERT
and MT-DNN, SWRs
still improve BERT test accuracy on 3 of 4 test sets and
MT-DNN test accuracy on 2 of 4 test sets.

\section{Analysis}
\label{sec:analysis}

In this section, we perform an ablation study and probe our models
with a diagnostic evaluation dataset (HANS).

\subsection{Do Our Models Actually Use SWRs?}

The DA models most consistently benefit from SWRs;
one question is if they are simply taking advantage of the increased
number of parameters. To examine this, we ablate the DA\texttt{+LF} and DA\texttt{+SA} models with
random noise in place of
the SWRs (denoted $\texttt{+LF}_\mathcal{N}$ and 
$\texttt{+SA}_\mathcal{N}$). In these experiments, $\mathbf{s} \sim
\mathcal{N}(0,\,1)$. 

First, we observe that the test accuracies of DA$\texttt{+LF}_\mathcal{N}$ and
DA$\texttt{+SA}_\mathcal{N}$ are less than or equal to their pretrained counterparts
(DA\texttt{+LF}, DA\texttt{+SA}) on all datasets (Table~\ref{table:nli-abl}).
DA\texttt{+SA} is harmed most by random
SWRs since \texttt{+SA} models directly use untransformed
SWRs when calculating attention; that is, \texttt{+SA} models cannot learn to
ignore the random noise.
 
\begin{table}[!tbp]
\centering
{\small
\begin{tabular}{l|cc|cc}
\toprule
Dataset & \texttt{+LF}   & $\texttt{+LF}_\mathcal{N}$ & \texttt{+SA} & $\texttt{+SA}_\mathcal{N}$ \\
\midrule
SNLI    & 84.8 & 84.2 (-0.6) & 83.2 & 75.6 (-7.6) \\
SciTail & 78.0 & 75.2 (-2.8) & 78.2 & 75.1 (-3.1) \\
MNLI-m  & 71.6 & 71.2 (-0.4) & 69.4 & 63.0 (-6.4) \\
MNLI-mm & 71.0 & 71.0 (-0.0) & 69.6 & 62.9 (-6.7) \\
\bottomrule
\end{tabular}
}
\caption{DA ablation study test accuracies.}
\label{table:nli-abl}
\end{table}

\begin{table}[!tbp]
\centering
{\small
\begin{tabular}{l|ccc}
\toprule
Dataset & DA   & DA$\texttt{+LF}_\mathcal{N}$ & DA$\texttt{+SA}_\mathcal{N}$ \\
\midrule
SNLI    & 82.1 & 84.2 (+2.1)       & 75.6 (-6.5)        \\
SciTail & 75.1 & 75.2 (+0.1)       & 75.1 (-0.0)        \\
MNLI-m  & 68.6 & 71.2 (+2.6)       & 63.0 (-5.6)        \\
MNLI-mm & 68.9 & 71.0 (+2.1)       & 62.9 (-6.0)        \\
\bottomrule
\end{tabular}
}
\caption{Test accuracies with random noise SWRs.}
\label{table:nli-abl-comp-da}
\end{table}


\begin{table*}[!tbp]
\centering
{\small
\begin{tabular}{lcccccccc}
\toprule
               & \multicolumn{2}{c}{SNLI} & \multicolumn{2}{c}{SciTail} & \multicolumn{2}{c}{MNLI-m} & \multicolumn{2}{c}{MNLI-mm} \\
                 \cmidrule(lr){2-3}\cmidrule(lr){4-5}\cmidrule(lr){6-7}\cmidrule(lr){8-9}
Model          & dev         & test        & dev         & test        & dev         & test        & dev         & test        \\
\midrule
DA                           & 81.7 & 82.1 & 78.3 & 75.1 & 68.6 & 68.6 & 69.4 & 68.9 \\
DA\texttt{+LF}               & 84.9 & 84.8 & 79.3 & 78.0 & 71.6 & 71.6 & 72.4 & 71.0 \\
DA$\texttt{+LF}_\mathcal{N}$ & 84.8 & 84.2 & 79.1 & 75.2 & 71.5 & 71.2 & 72.3 & 71.0 \\
DA\texttt{+SA}               & 83.1 & 83.2 & 82.8 & 78.2 & 69.8 & 69.4 & 71.0 & 69.6 \\
DA$\texttt{+SA}_\mathcal{N}$ & 75.6 & 75.6 & 69.9 & 75.1 & 63.7 & 63.0 & 63.7 & 62.9 \\
\bottomrule
\end{tabular}
}
\caption{DA ablation study dev and test accuracies.}
\label{table:nli-abl-full}
\end{table*}

We also observe that DA$\texttt{+LF}_\mathcal{N}$ performs \emph{better} than DA
on every dataset, while DA$\texttt{+SA}_\mathcal{N}$ performs \emph{worse} than
DA on every dataset except SciTail, where the test accuracy is unaffected
(Table~\ref{table:nli-abl-comp-da}).
This is consistent with our previous observation that \texttt{+SA} models
cannot learn to ignore random noise while \texttt{+LF} models can.
While it is surprising that DA$\texttt{+LF}_\mathcal{N}$ consistently
improves performance over DA, we suspect this gain is the result of additional model parameters.
We report our full ablation study results in Table~\ref{table:nli-abl-full}.

\subsection{Why Do SWRs Improve NLI?}

The Heuristic Analysis for NLI Systems dataset (HANS; \citealp{mccoy2019right})
is a controlled evaluation set for probing
whether an NLI model learns fallible syntactic heuristics.
HANS (30k pairs, labeled either \emph{entailment} and \emph{non-entailment})
is generated with templates that ensure the premise
and hypothesis exhibit special syntactic relationships. The dataset
is also balanced (i.e., random guessing should score around 50\%).

\citet{mccoy2019right} show that NLI models learn heuristics around
lexical overlap, subsequence relationships, and syntactic constituency between the
premise and hypothesis;\footnote{These syntactic
properties are commonly found in NLI datasets.} they report that four NLI models trained on MNLI
(including DA, ESIM, and BERT)
score $\sim$50\% overall and close to 0\% on the \emph{non-entailment} label.

\begin{table*}[tb]
\centering
{\small
\begin{tabular}{lcccccccc}
\toprule
      & & \multicolumn{3}{c}{Correct: \emph{Entailment}} & \multicolumn{3}{c}{Correct: \emph{Non-entailment}} \\
        \cmidrule(lr){3-5}\cmidrule(lr){6-8}
Model & Train Data & Lexical & Subseq. & Const. & Lexical & Subseq. & Const. & Avg. \\
\midrule
DA & SciTail             & 93.4 & 89.1 & 90.3 & 7.1  & 9.9  & 11.0 & 50.1 \\
DA\texttt{+SA} & SciTail & 88.3 & 91.3 & 81.1 & 20.1 & 19.2 & 20.5 & 53.4 \\
\midrule
ESIM & SciTail         & 9.6  & 11.0 & 14.6 & 93.8 & 91.1 & 87.5 & 51.3 \\
ESIM\texttt{+SA} & SciTail & 78.0 & 84.1 & 84.4 & 29.2 & 19.2 & 16.8 & 52.0 \\
\midrule
BERT & SciTail             & 100  & 100  & 99.1 & 0.1 & 0.6  & 1.3 & 50.2 \\
BERT\texttt{+LF} & SciTail & 98.7 & 99.4 & 94.2 & 7.0 & 7.0 & 13.6 & 53.2 \\
\midrule
MT-DNN & MNLI-mm             & 99.0 & 100  & 99.8 & 53.1 & 2.8 & 3.8 & 59.7 \\
MT-DNN\texttt{+LF} & MNLI-mm & 99.0 & 99.8 & 99.8 & 58.3 & 3.8 & 5.2 & 61.0 \\
\bottomrule
\end{tabular}
}
\caption{Model accuracies on HANS.}
\label{table:hans-top}
\end{table*}

We evaluate a subset of our models on HANS.
In particular, we choose the model/dataset combinations whose test accuracies
improve the most
with SWRs: DA\texttt{+SA} on SciTail (+3.1\%), ESIM\texttt{+SA} on SciTail (+5.1\%),
BERT\texttt{+LF} on SciTail (+2.0\%), and MT-DNN\texttt{+LF} on
MNLI-mm (+0.5\%).
We find SWRs improve overall accuracy on HANS (1--3\%) over their non-SWR counterparts
for all four models listed above.
We also observe that SWRs significantly reduce the models' reliance on 
fallible syntactic heuristics---that is, the accuracies between the \emph{entailment}
and \emph{non-entailment} labels are far more balanced.
Our full experimental results on HANS can be found in Table~\ref{table:hans-top}.

\section{Related Work}
There is a long line of work on incorporating syntactic features for semantic
tasks, including NLI. Some earlier approaches include learning syntactic rules
indicative of entailment (e.g., \citealp{barhaim2007, mehdad2010}) or features
derived from tree transformations between premise and hypothesis
\citep{heilman2010, wang2010}. More recent research has incorporated syntactic
structures into neural architectures (e.g., tree LSTMs in
\citealt{chen2016enhanced}; graph-based architecture in \citealt{Khot2018-th}).

Other recent work has experimented with incorporating syntactic learning
objectives.  \citet{strubell2018linguistically} and
\citet{swayamdipta2018syntactic} used multitask learning of syntactic and semantic
tasks to transfer syntactic features, improving performance on semantic role
labeling and, in the latter case, coreference resolution.

\section{Conclusion}
In this paper, we introduced a simple and broadly applicable method for incorporating syntax into NLI models, through the use of contextual vector representations from a pretrained parser. We demonstrated that our method often improves accuracy for four NLI models (DA, ESIM, BERT, MT-DNN) across three standard NLI datasets. Our findings demonstrate the
usefulness of
syntactic information in semantic models and motivate future research into
syntactically informed models.

\section*{Acknowledgments}

We thank Joshua Bean, Lynsey Liu, and Aaron Johnston for their help in conducting preliminary experiments. We are also grateful to members of the UW NLP community and anonymous reviewers for their comments and suggestions. This research was supported in part by a NSF Graduate Fellowship to LHL.


\bibliographystyle{acl_natbib}
\bibliography{references}
\clearpage

\onecolumn

\begin{appendices}

\section{Hyperparameters and Tuning Procedure} \label{appendix:hyperparameters}

In this section, we report our hyperparameters and describe our tuning procedure.

\subsection{Word Embeddings}

The parser uses 100-dimensional
GloVe word embeddings \citep{pennington2014glove} trained on
Wikipedia/Gigaword while
the rest of our models use 300-dimensional GloVe word
embeddings.

\subsection{Optimizer}

We train DA, ESIM, and BERT with Adam
\citep{kingma2014adam}, and for  MT-DNN, we use the Adamax variant of Adam following
\citet{liu2019multi}.

\subsection{Decomposable Attention Model}

For the DA baseline and each variant, we randomly sample 30 configurations
and choose the best performing dev.~configuration on SciTail.
We sometimes round the hyperparameters for ease of implementation.
We use the hyperparameters chosen on SciTail for all other datasets.
Each hyperparameter is sampled from a range, either uniformly
or log-uniformly (Table~\ref{tab:da-tune}).
Our chosen DA hyperparameters are reported in Table~\ref{tab:da-params}.

\begin{table}[!htbp]
\centering
{\small
\begin{tabular}{lcc}
\toprule
Param. & Range & Sampling Method \\
\midrule
Learning Rate   & [\num{1e-6}, 0] & Log-Uniform \\
Att. FF HD    & [100, 300]      & Uniform \\
Comp. FF HD   & [100, 400]      & Uniform \\
Agg. FF HD & [100, 400]      & Uniform \\
Att. FF Dropout & [0.2, 0.7]    & Uniform \\
Comp. FF Dropout & [0.2, 0.7]   & Uniform \\
Agg. FF Dropout & [0.2, 0.7] & Uniform \\
\bottomrule
\end{tabular}
}
\caption{DA hyperparameter sampling ranges and methods. FF stands for feed-forward and HD stands for hidden dimensions.}
\label{tab:da-tune}
\end{table}

\begin{table}[!htbp]
\centering
{\small
\begin{tabular}{lccc}
\toprule
Param. & DA & DA\texttt{+LF} & DA\texttt{+SA} \\
\midrule
Learning Rate   & \num{3e-4} & \num{3e-4} & \num{3e-4} \\
Att. FF HD    & 295 & 250 & 295 \\
Comp. FF HD   & 108 & 400 & 108      \\
Agg. FF HD & 172 & 150 & 295      \\
Att. FF Dropout & 0.29  & 0.3 & 0.29  \\
Comp. FF Dropout & 0.34 & 0.3 & 0.34  \\
Agg. FF Dropout & 0.54 & 0.3 & 0.54 \\
\bottomrule
\end{tabular}
}
\caption{Chosen DA hyperparameters. FF stands for feed-forward and HD stands for hidden dimensions.}
\label{tab:da-params}
\end{table}

\newpage

\begin{table}[!htbp]
\centering
{\small
\begin{tabular}{lcc}
\toprule
Param. & Range & Sampling Method \\
\midrule
Learning Rate  & [\num{1e-4}, -1] & Log-Uniform \\
Model Dropout  & [0.2, 0.7] & Uniform \\ 
Output FF Dropout & [0.2, 0.7] & Uniform \\
\bottomrule
\end{tabular}
}
\caption{ESIM hyperparameter sampling ranges and methods. FF stands for feed-forward.}
\label{tab:esim-tune}
\end{table}

\begin{table}[!htbp]
\centering
{\small
\begin{tabular}{lccc}
\toprule
Param. & DA & DA\texttt{+LF} & DA\texttt{+SA} \\
\midrule
Learning Rate   & \num{3e-4} & \num{3e-4} & \num{3e-4} \\
Att. FF HD    & 295 & 250 & 295 \\
Comp. FF HD   & 108 & 400 & 108      \\
Agg. FF HD & 172 & 150 & 295      \\
Att. FF Dropout & 0.29  & 0.3 & 0.29  \\
Comp. FF Dropout & 0.34 & 0.3 & 0.34  \\
Agg. FF Dropout & 0.54 & 0.3 & 0.54 \\
\bottomrule
\end{tabular}
}
\caption{Chosen ESIM hyperparameters. FF stands for feed-forward.}
\label{tab:esim-params}
\end{table}

\subsection{ESIM}

For the ESIM baseline and each variant, we randomly sample 10 configurations (Tables~\ref{tab:esim-tune}, \ref{tab:esim-params}).

\subsection{BERT and MT-DNN}

We use the hyperparameters reported by \citet{Devlin:19} and \citet{liu2019multi}.
The only exception is that for MT-DNN\texttt{+LF}, we find that in some cases
a longer warmup (0.3) improves performance.

\end{appendices}

\end{document}